\documentclass{article}

\usepackage{microtype}
\usepackage{graphicx}
\usepackage{booktabs}
\usepackage{enumitem}
\usepackage{float}
\usepackage{placeins}
\usepackage{wrapfig}
\usepackage{hyperref}
\usepackage{tikz}
\usetikzlibrary{arrows.meta,positioning,fit,backgrounds}


\usepackage[accepted]{icml2025}
\makeatletter
\renewcommand{\ICML@appearing}{Preprint. Copyright 2026 by the author(s).}
\renewcommand{\printAffiliationsAndNotice}[1]{%
\stepcounter{@affiliationcounter}%
{\let\thefootnote\relax\footnotetext{\hspace*{-\footnotesep}#1%
\forloop{@affilnum}{1}{\value{@affilnum} < \value{@affiliationcounter}}{
\textsuperscript{\arabic{@affilnum}}\csname @affilname\the@affilnum\endcsname%
}.

\ \\
\Notice@String
}
}
}
\makeatother

\usepackage{xcolor}
\usepackage{colortbl}
\usepackage{multirow}
\usepackage{rotating}
\usepackage{amsmath}
\usepackage{amssymb}
\usepackage{mathtools}

\newcommand{\method}{ActionST}
\newcommand{\Xh}{\mathbf{X}^{\mathrm{h}}}
\newcommand{\Xz}{\mathbf{X}^{0}}
\newcommand{\XI}{\mathbf{X}^{I}}
\newcommand{\hXz}{\hat{\mathbf{X}}^{0}}
\newcommand{\hXI}{\hat{\mathbf{X}}^{I}}

\newcommand{\pmcell}[2]{#1 {\tiny$\pm$ #2}}
\newcommand{\bestpm}[2]{\textbf{#1} {\tiny$\pm$ #2}}
\newcommand{\secondpm}[2]{\underline{#1} {\tiny$\pm$ #2}}

\usepackage[capitalize,noabbrev]{cleveref}

\icmltitlerunning{Action-Conditioned Spatio-Temporal Forecasting}

\begin{document}

\twocolumn[
\icmltitle{Action-Conditioned Spatio-Temporal Forecasting}

\icmlsetsymbol{equal}{*}
\begin{icmlauthorlist}
\icmlauthor{Yutong Xia}{nus}
\icmlauthor{Yangyou Liu}{scu}
\icmlauthor{Yuxuan Liang}{hkustgz}
\icmlauthor{Yuankai Wu}{scu}
\icmlauthor{See-Kiong Ng}{nus}
\icmlauthor{Roger Zimmermann}{nus}
\end{icmlauthorlist}
\icmlaffiliation{nus}{National University of Singapore}
\icmlaffiliation{scu}{Sichuan University}
\icmlaffiliation{hkustgz}{The Hong Kong University of Science and Technology (Guangzhou)}
\icmlkeywords{spatio-temporal forecasting, action-conditioned forecasting, intervention forecasting, reference-and-response prediction, paired simulation}
\vskip 0.3in
]

\printAffiliationsAndNotice{}

\begin{abstract}
\vspace{1em}
We study spatio-temporal forecasting in networked systems under specified actions.
The challenge is to capture the action's effects without predicting spurious changes in unaffected regions.
However, direct supervision and evaluation of these effects require paired futures with and without the action, whereas observational trajectories reveal only the realized future.
We introduce \textbf{PairST}, a benchmark comprising six datasets across traffic, power transmission and water distribution, with paired reference and intervened futures simulated from the same system state.
These pairs enable separate evaluation of affected-region forecast accuracy and quiet-region response error.
We further propose \textbf{\method{}}, which forecasts the intervened future by writing an explicit response onto an intervention-independent reference forecast.
Specifically, node--time addressing constructs a query for each response position, a state-conditioned response writer computes candidate responses, and a graph-aware locality gate controls their strength.
Experiments on six datasets demonstrate the effectiveness of \method{} at new action targets, with the lowest mean absolute error on five datasets and the best average rank.
Our source code and the PairST benchmark are available at \url{https://github.com/yutong-xia/ActionST}.
\end{abstract}

\section{Introduction}
\label{sec:intro}

Spatio-temporal (ST) forecasting plays an important role in decision-making across many domains, such as traffic~\citep{li2018dcrnn,wu2019graphwavenet}, energy~\citep{enhancedspatiote2022,niu2025solar}, air quality~\citep{yi2018distributed,liang2023airformer} and public health~\citep{deng2020colagnn,panagopoulos2021transfer}.
Many of these decisions involve taking \emph{actions} to improve system operation, such as adjusting traffic signals~\citep{wei2019colight,lai2025llmlight,zou2026trafficr1} or disconnecting a power line~\citep{fisher2008switching,marot2021l2rpn}.
To inform these decisions, we therefore ask: \textit{how will signals across the network evolve under a specified action?}
To study this question systematically, we formulate the task of \emph{action-conditioned spatio-temporal forecasting}.

\begin{figure}[t]
\centering
\vspace{1em} %
\includegraphics[width=\columnwidth]{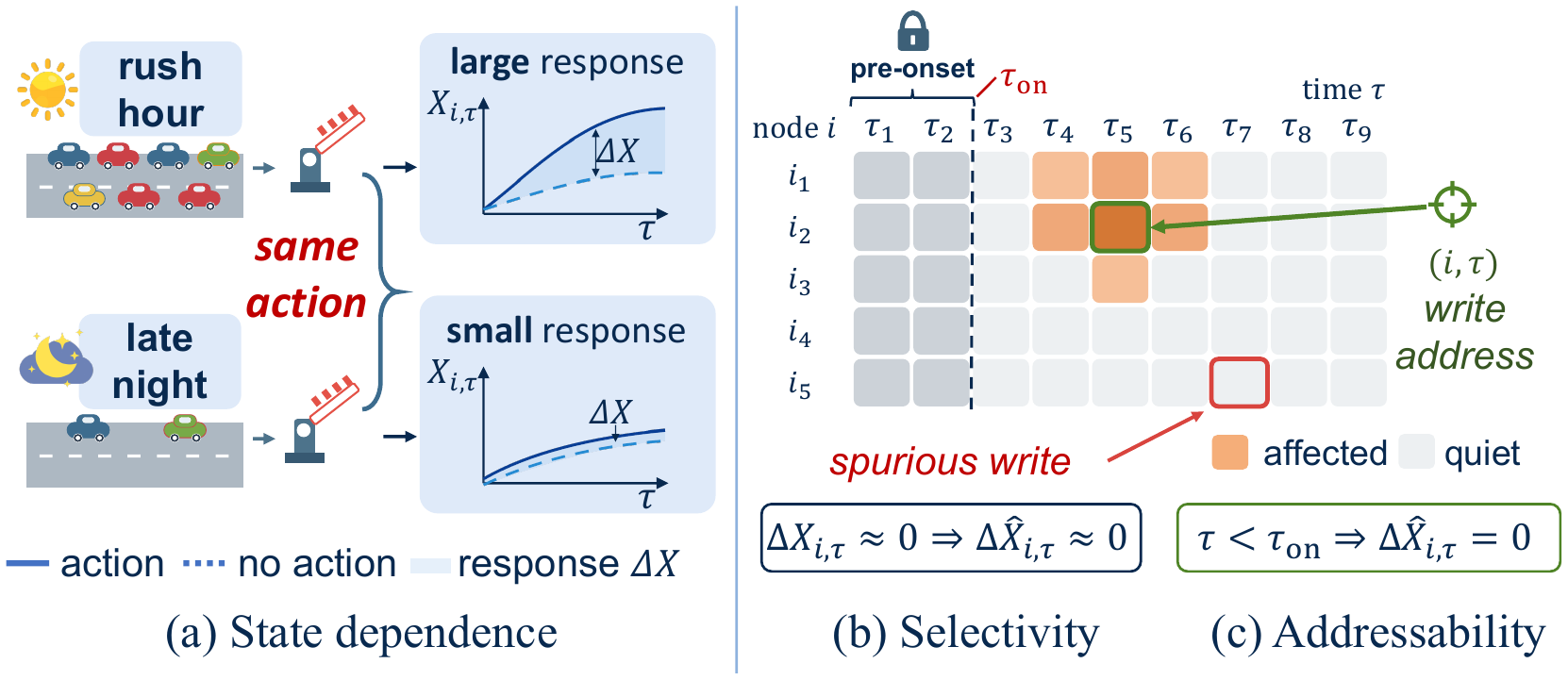}
\caption{Three requirements for action-conditioned spatio-temporal forecasting via response modeling.}
\label{fig:motivation}
\vspace{-1em}
\end{figure}

To evaluate whether forecasts capture the changes caused by an action, we seek a benchmark that pairs futures with and without it from the same system state.
However, existing benchmarks~\citep{li2018dcrnn,liu2023largest,gou2025traffident,li2025urbanev} provide histories and observed futures, but lack paired futures with and without a specified action.
In practice, real-world systems cannot generally be reset to observe both futures from the same state.
Large-scale field experiments are also costly and challenging to implement~\citep{baumsnow2015causal}, with additional ethical constraints~\citep{hernan2020whatif}.

Beyond the data gap, this task also calls for models that capture how specified actions affect a network over time.
Conventional ST forecasters do not explicitly model actions~\citep{yu2018stgcn,wu2019graphwavenet,bai2020agcrn}.
Event-conditioned forecasters additionally incorporate observed incidents into prediction~\citep{wang2026conformer,fan2026igstgnn,kitahara2026castanet}.
Meanwhile, action-conditioned world models address domain-specific settings, including antenna adjustments~\citep{qi2026mobiwm}, policy changes~\citep{ma2026policy4ood} and robot motion~\citep{chen2026flowwam}.
Here, we study a common forecasting formulation for operational actions across different physical networks.

Therefore, to address the data gap, we construct a benchmark termed \underline{\textbf{Pair}}ed Intervention Futures for \underline{\textbf{S}}patio-\underline{\textbf{T}}emporal Systems (\textbf{PairST}), which simulates futures without and with the action from the same system state, termed the \emph{reference future} and \emph{intervened future}, respectively.
PairST comprises six datasets across traffic, power transmission and water distribution.
Its paired futures also support direct supervision and evaluation of action-induced changes.

To forecast the intervened future, we model the action-induced change explicitly as a \emph{response} relative to the reference future.
Paired futures make this response observable, yet predicting it poses three requirements: \textit{state dependence}, \textit{selectivity}, and \textit{addressability}.
First, the response should depend on the pre-intervention system state (Figure~\ref{fig:motivation}a).
Closing the same road at rush hour and late at night can have very different effects.
Second, the predicted response should capture changes at affected node--time positions while remaining near zero in quiet regions (Figure~\ref{fig:motivation}b).
Third, the response should be controllable at individual node--time positions, so that temporal constraints, e.g., no response before onset, can be enforced locally (Figure~\ref{fig:motivation}c).

To meet these three requirements, we therefore propose a novel framework termed \underline{\textbf{Action}}-conditioned \underline{\textbf{S}}patio-\underline{\textbf{T}}emporal Neural Network (\textbf{\method{}}), which forecasts the intervened future by writing an explicit response onto an intervention-independent reference forecast.
Specifically, node--time addressing constructs a query for each response position, a state-conditioned response writer computes candidate responses, and a graph-aware locality gate controls their strength.
The response path also keeps the response exactly zero under an empty intervention and before onset, and lets the response at each future step depend only on the intervention executed by then.

Our contributions are summarized as follows.
\begin{itemize}[leftmargin=*,topsep=2pt,itemsep=2pt,parsep=0pt]
\item
\textbf{A Paired Multi-System Benchmark.} We build PairST, a benchmark with six datasets across three different domains, including traffic, power transmission and water distribution.
It provides paired futures with and without an action from the same system state, supporting direct supervision and evaluation of action-induced changes.
\item
\textbf{An Action-Conditioned Forecasting Model.} We propose \method{}, which writes gated, state-conditioned responses at individual node--time positions.
The framework guarantees null identity, pre-onset invariance and prefix consistency by construction.
\item
\textbf{Comprehensive Experiments.} We conduct comprehensive experiments on six PairST datasets against seven baselines, where \method{} achieves the lowest mean absolute error (MAE) on five datasets and the best average rank.
Further analyses evaluate response propagation, component contributions and hyperparameter sensitivity.
\end{itemize}

\section{Related Work}
\label{sec:related}

\textbf{Spatio-Temporal Forecasting.}
Spatio-Temporal (ST) forecasters capture network dependencies and temporal patterns using graph convolutions~\citep{li2018dcrnn,yu2018stgcn,wu2019graphwavenet,bai2020agcrn}, node and time embeddings~\citep{liu2023staeformer,shao2022stid} or diffusion~\citep{wen2023diffstg}.
OpenCity~\citep{li2024opencity} and UniST~\citep{yuan2024unist} learn transferable representations through pretraining.
Event-conditioned methods, including ConFormer~\citep{wang2026conformer}, IGSTGNN~\citep{fan2026igstgnn} and CASTANET~\citep{kitahara2026castanet}, additionally incorporate observed incidents.
In this work, we consider actions specified at prediction time, including actions never taken from the current state.

\textbf{Action-Conditioned Models.}
Action-conditioned models predict futures under antenna adjustments in mobile networks~\citep{qi2026mobiwm}, policy interventions~\citep{ma2026policy4ood} or optical-flow actions for video generation~\citep{chen2026flowwam}.
Related designs include LEAP nets' latent perturbations that vanish for the reference topology~\citep{donon2020leap} and internal interventions that redirect video predictions~\citep{wang2026causalwritability}.
In contrast, \method{} combines an intervention-independent reference with gated, state-conditioned response writing at node--time positions.

\textbf{Interventional and Counterfactual Prediction.}
Another line of work estimates outcomes under alternative treatment sequences from observational data.
CRN~\citep{bica2020crn} and Causal Transformer~\citep{melnychuk2022causaltf} address time-varying confounding, while GST-UNet~\citep{oprescu2025gstunet} also considers ST interventions and interference.
In contrast, PairST provides paired simulator futures for directly supervising and evaluating action-induced changes.

\begin{table*}[t]
\caption{Statistics of PairST datasets. Units: intersections for traffic, transmission lines for power and links for water. $a_m$: active-response threshold of Eq.~\ref{eq:response-regions}, in vehicles for traffic, percent of the thermal limit for power and L/s for water. Locations, Days, Pairs: training/validation/test counts.}
\label{tab:data}
\centering
\footnotesize
\setlength{\tabcolsep}{3pt}
\renewcommand{\arraystretch}{1.15}
\begin{tabular*}{\textwidth}{@{\extracolsep{\fill}}lllrcccc@{}}
\toprule[1.5pt]
\textbf{System} & \textbf{Intervention} & \textbf{Dataset} & \textbf{Units} & $\boldsymbol{a_m}$ & \textbf{Locations} & \textbf{Days} & \textbf{Pairs} \\
\midrule
\multirow{2}{*}{Traffic (CityFlow)} & \multirow{2}{*}{signal-phase change} & PairST-T16 & 16 & 4 & 10/3/3 & 125/25/50 & 10{,}000/600/1{,}200 \\
 & & PairST-T196 & 196 & 4 & 16/4/4 & 80/16/32 & 10{,}240/512/1{,}024 \\
\midrule
\multirow{2}{*}{Power (Grid2Op)} & \multirow{2}{*}{line disconnection} & PairST-P59 & 59 & 7.8 & \multirow{2}{*}{16/4/4} & 75/20/22 & 9{,}600/639/704 \\
 & & PairST-P186 & 186 & 7.4 & & 80/16/32 & 10{,}236/512/1{,}024 \\
\midrule
\multirow{2}{*}{Water (EPANET)} & \multirow{2}{*}{pipe closure} & PairST-W119 & 119 & 10.2 & 16/4/4 & 80/16/32 & 10{,}240/512/1{,}024 \\
 & & PairST-W40 & 40 & 0.466 & 13/3/3 & 96/20/40 & 9{,}984/480/960 \\
\bottomrule[1.5pt]
\end{tabular*}
\end{table*}

\section{Problem Formulation}
\label{sec:problem}

Action-conditioned ST forecasting predicts future network signals given an observed history and a specified action.

\textbf{Network Signals.}
Let $G=(V,E)$ denote a network with $N=|V|$ nodes, and let $\Xh\in\mathbb{R}^{N\times T\times D}$ contain $D$ observed features over the $T$ steps before the forecast time.
We predict $D'$ output features over an $S$-step future window, indexed by $\tau\in\{1,\ldots,S\}$.

\textbf{Actions.}
An \emph{action} is an operation applied to a specified component of the network over a specified time interval.
We represent it by a structured intervention record
\[
I=(\kappa,i^\star,\tau_{\mathrm{on}},\ell,\eta),
\]
where $\kappa$ denotes the operation type, $i^\star\in V$ its target, $\tau_{\mathrm{on}}$ its onset position in the forecast window, $\ell>0$ its duration in time steps, and $\eta\in\mathcal{P}_\kappa$ its parameters from the type-specific parameter space $\mathcal{P}_\kappa$.

In a traffic network, for example, an action $I$ can hold intersection $i^\star$ at signal phase $\eta$ for $\ell$ steps starting at $\tau_{\mathrm{on}}$, where $\kappa$ is this phase-holding operation, as a traffic controller does to manage traffic around a major event such as a marathon.
Analogous actions switch a transmission line out of service for maintenance or close a valve to isolate a water pipe for repair.
Such an action targets one component for a limited time, yet its effects can spread to neighboring components and persist after normal operation resumes.
The empty record $\varnothing$ denotes no action.

\textbf{Prediction Task.}
Given the historical observations $\Xh$, the network graph $G$ and the action record $I$, we predict the \emph{intervened future} $\XI\in\mathbb{R}^{N\times S\times D'}$, namely how the network signals evolve under the action.
\textbf{Our goal} is to learn a function $\mathcal{F}_{\theta}(\cdot)$ that maps these inputs to the intervened future:
\begin{equation}
[\Xh,G,I]\xrightarrow{\mathcal{F}_{\theta}(\cdot)}[\hXI].
\label{eq:problem}
\end{equation}

\textbf{Action-Induced Response.}
To characterize the changes caused by the action, we compare the intervened future $\XI$ with the \emph{reference future} $\Xz$, the future from the same initial state without the action, and define the \emph{response} $\Delta\mathbf{X}=\XI-\Xz$.
This difference separates action-induced changes from the system's evolution without the action.

\section{\texorpdfstring{PairST}{PairST} Benchmark}
\label{sec:pairst}
To provide a benchmark for action-conditioned ST forecasting, we build PairST with simulators of traffic, power transmission and water distribution networks.
For each system state and action, the simulator generates the reference and intervened futures from the same state.

\subsection{Dataset Construction}
\label{sec:pairst-systems}
PairST contains six datasets across three different domains, with two datasets per domain.
Each dataset is one physical network named by its system and number of graph units.
Specifically, the three domains are described below:
\begin{itemize}[leftmargin=*,topsep=2pt,itemsep=2pt,parsep=0pt,before={\interlinepenalty=10000}]
\item \textbf{Traffic.} We generate PairST-T16 and PairST-T196 in CityFlow~\citep{zhang2019cityflow} on CoLight's Hangzhou and New York roadnets~\citep{wei2019colight}.
An intervention changes the signal phase of one intersection, and the forecast variable is the vehicle count.
\item \textbf{Power.} We generate PairST-P59 and PairST-P186 in Grid2Op's NeurIPS-2020 and WCCI-2022 environments~\citep{donnot2020grid2op,marot2021l2rpn,serre2022carbon}.
An intervention temporarily disconnects one transmission line, and the forecast variable is line loading in percent of the thermal limit.
\item \textbf{Water.} We generate PairST-W119 and PairST-W40 on EPANET's~\citep{rossman2020epanet} Net3 and Net2 networks.
An intervention temporarily closes one pipe, and the forecast variable is signed link flow in litres per second.
\end{itemize}
Table~\ref{tab:data} summarizes the network sizes, action types and data splits.
Appendix~\ref{app:pairst-details} gives the graph of each dataset (Figure~\ref{fig:pairst-examples}c) and the complete simulator, record and qualification settings.

\textbf{Data Splits.}
To test generalization across target locations, PairST evaluates forecasts at \emph{new action targets}, namely locations never acted on in training.
We therefore split each dataset by action target, so that validation and test actions act only at locations never targeted in training, and keep training, validation and test days disjoint so that no system state is shared across splits.
All splits use the same physical network and action settings (action types, onsets and durations).

\subsection{Dataset Analysis}
\label{sec:pairst-response}
We analyze the sparsity and spatial extent of action-induced changes in PairST.
To quantify these properties, we first distinguish affected and quiet positions using the paired response.

\textbf{Response Regions.}
An action can substantially change some network signals while leaving others nearly unchanged.
A \emph{position} is indexed by a node, a future step and an output feature, written $u=(i,\tau,c)$ with $i\in V$, $\tau\in\{1,\ldots,S\}$ and $c\in\{1,\ldots,D'\}$, and $\Omega$ contains all positions in the forecast window.
For each dataset $m$, we fix an active-response threshold $a_m$ in its native unit before model fitting and define
\begin{equation}
\begin{aligned}
\Omega^{\mathrm{aff}}_m&=\{u\in\Omega:|\Delta X_u|\geq a_m\},\\
\Omega^{\mathrm{quiet}}_m&=\{u\in\Omega:|\Delta X_u|<a_m/2\}.
\end{aligned}
\label{eq:response-regions}
\end{equation}
Positions between the two thresholds remain in the overall evaluation but enter neither region-specific metric.
On the traffic datasets $a_m$ is four vehicles, and on power and water it is half the median peak $|\Delta X|$ outside the target over training pairs (Table~\ref{tab:data}).

\textbf{Sparsity.}
Over training pairs, responses affect only a small fraction of the forecast window.
Affected positions account for only 0.5\%--7.3\% of a dataset's output positions, while quiet positions account for 90\%--99\% (Appendix Table~\ref{tab:pairst-response}).
This sparsity follows from the actions themselves: each changes one component for a limited time, so most nodes and steps remain close to the reference future.

\begin{figure}[!t]
\centering
\includegraphics[width=\columnwidth]{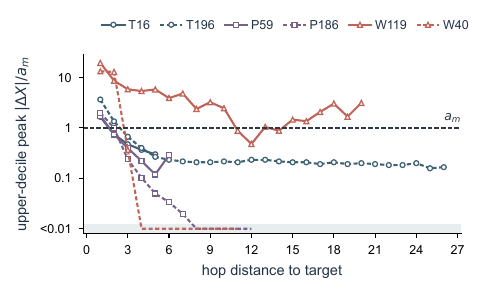}%
\vspace{-1em}
\caption{Spatial extent of the simulator response. 90th percentile of the window peak $|\Delta X|$ across non-target units at each hop distance from the target, relative to $a_m$, on training pairs. Hops reached from fewer than half of the training target locations are omitted, and values below $0.01\,a_m$ lie in the grey band.}
\label{fig:pairst-reach}
\end{figure}

\textbf{Spatial Extent.}
Responses differ markedly across systems in how far they spread from the target.
On five datasets, even the most responsive tenth of non-target units reaches $a_m$ only within two hops of the target, whereas on PairST-W119 it stays near or above $a_m$ out to 20 hops (Figure~\ref{fig:pairst-reach}).

Together, these properties call for models that confine predicted changes to affected positions while adapting their spatial reach to each system.

\section{Methodology}
\label{sec:method}
\begin{figure*}[!t]
\centering
\vspace{-0.5em}
\includegraphics[width=\textwidth]{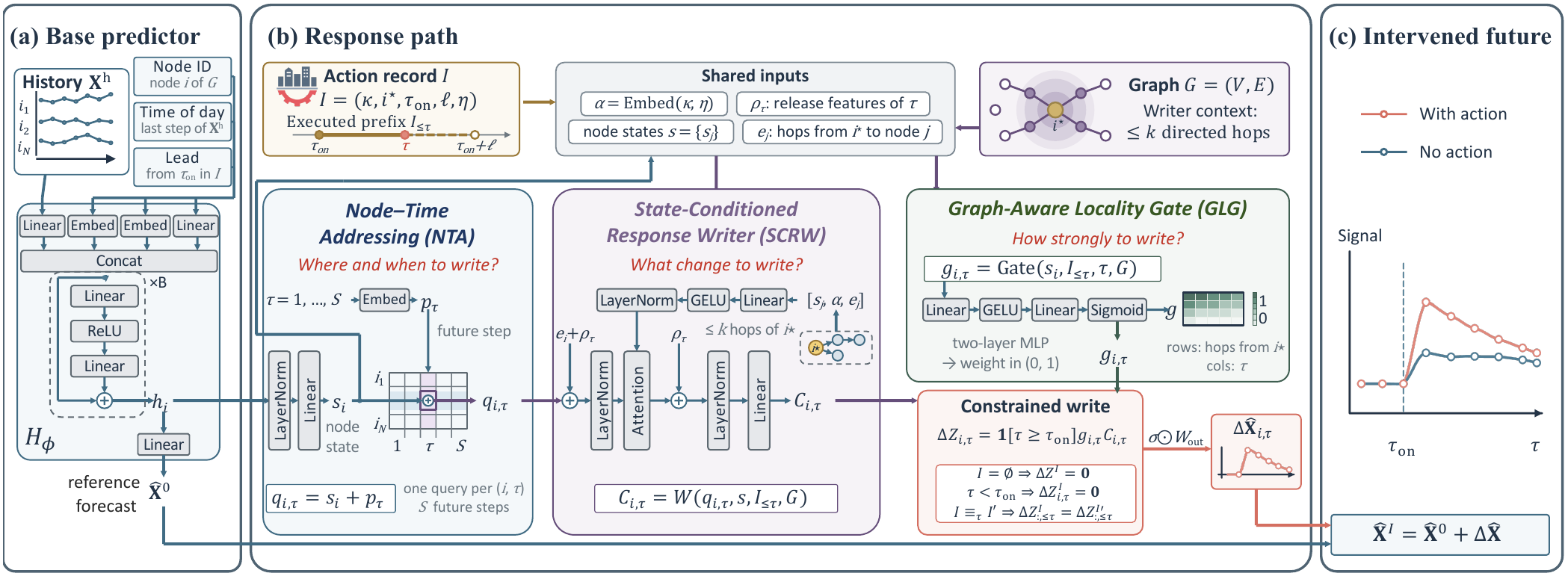}
\vspace{-2em}
\caption{The framework of \method{}.}
\label{fig:framework}
\vspace{-0.5em}
\end{figure*}

We propose \method{} to meet the requirements of addressability, state dependence and selectivity (Figure~\ref{fig:motivation}).

\textbf{Framework Overview} (Figure~\ref{fig:framework}).
\method{} first forecasts the future network signals as if no action were taken, then predicts how the action changes this forecast.
A base predictor $H_\phi$ maps the historical observations $\Xh$ to two outputs, i.e., a hidden vector $h_i\in\mathbb{R}^{d_h}$ for every node $i$ and the reference forecast $\hXz$ (Figure~\ref{fig:framework}a).
To distinguish nodes and times whose recent histories look alike, it also takes the node identity and the time of day as inputs, following STID~\citep{shao2022stid}, together with the forecast lead, i.e., the time from the forecast time to the onset, which places the output window in time.
Here $h_i$ is the pre-intervention representation of node $i$ at forecast time, which the response path takes as input, and $\hXz$ is the forecast to which the predicted change is added at the end.

A response path then predicts the changes caused by the action (Figure~\ref{fig:framework}b) through three modules: \textbf{Node--Time Addressing (NTA)} constructs a query for each node and future step, the \textbf{State-Conditioned Response Writer (SCRW)} computes a candidate response at each query, and the \textbf{Graph-Aware Locality Gate (GLG)} controls how strongly each candidate is written.
Together, they determine the position, content and strength of each write in a response state $\Delta Z\in\mathbb{R}^{N\times S\times d}$, and a bias-free linear readout decodes it into a change that is added to the reference forecast (Figure~\ref{fig:framework}c).
The forecast at node $i$ and future step $\tau$ is
\begin{equation}
\hXI_{i,\tau}=\hXz_{i,\tau}+\boldsymbol{\sigma}\odot\big(W_{\mathrm{out}}\Delta Z_{i,\tau}\big),
\label{eq:reference-response}
\end{equation}
where $\hXI_{i,\tau},\hXz_{i,\tau}\in\mathbb{R}^{D'}$, $\Delta Z_{i,\tau}\in\mathbb{R}^{d}$ is the response state written at that node and step, $d$ is the width of the response path, $W_{\mathrm{out}}\in\mathbb{R}^{D'\times d}$ is the weight of the readout, $\boldsymbol{\sigma}\in\mathbb{R}^{D'}$ holds the training standard deviation of each output feature and $\odot$ is the element-wise product.
The layer-level architectures of the base predictor and the response path are given in Appendix~\ref{app:architecture}.
Next, we introduce the three modules of the response path in Figure~\ref{fig:framework}b, i.e., NTA, SCRW and GLG.

\subsection{Node--Time Addressing}
\label{sec:response}
To predict action-induced changes separately at each node and future step, the response path needs a distinct query for each node--time position $(i,\tau)$.
NTA builds this query from the pre-intervention state of node $i$ and the future step $\tau$ being predicted, which provides \textit{addressability}.

In Figure~\ref{fig:framework}b, a layer normalization followed by a linear map projects the node representation $h_i$ of the base predictor to a state $s_i\in\mathbb{R}^{d}$ of the response path, and a learned embedding $p_\tau\in\mathbb{R}^{d}$ identifies the output position $\tau$.
Their sum forms the query
\begin{equation}
q_{i,\tau}=s_i+p_\tau.
\end{equation}
Because the output window is anchored at the onset, $\tau$ also fixes the query's time relative to the intervention.
SCRW (Section~\ref{sec:writing}) then answers each query with a candidate response.

\subsection{State-Conditioned Response Writer}
\label{sec:writing}
The same action can have very different effects under different operating conditions, e.g., at rush hour and late at night (Section~\ref{sec:intro}).
To capture this \textit{state dependence}, SCRW answers each node--time query with a candidate response computed from the action and the pre-intervention states of the nodes around the target.

\textbf{Prefix-Causal Encoding.}
By a future step $\tau$, only the part of the action executed so far can have affected the network, so two interventions that differ only in duration should have the same predicted response until the shorter one ends, which matters in practice because the end of an action is often uncertain when the forecast is issued.
Inspired by the causal convolutions of WaveNet~\citep{oord2016wavenet}, we therefore encode the intervention so that no input of the response path reveals how long it lasts beyond each output position.
We write $I_{\le\tau}$ for the part of the record executed up to $\tau$, and $I\equiv_{\tau}I'$ if two records with the same forecast window induce the same executed action at every step up to $\tau$.
In place of the duration, the response path reads the \emph{release features} of $\tau$, namely the time since onset, the time since the intervention ended (zero while it is active), and indicators of whether it is active or has ended.

\textbf{State-Conditioned Attention.}
The query at $(i,\tau)$ asks how the action changes node $i$ at step $\tau$.
To place it relative to the action, the writer adds to $q_{i,\tau}$ a target-relative encoding $e_i\in\mathbb{R}^{d}$ of the hop distance and direction of node $i$ from the target and its out-degree, and a release encoding $\rho_\tau\in\mathbb{R}^{d}$ of the release features of $\tau$.
Because the writer identifies the target only through $e_i$, it treats new and training action targets in the same way.
The query then attends to the nodes within $k$ directed hops of the target, the target included, with $k$ set per dataset (Appendix~\ref{app:config}), whose keys and values project $[s_j,\alpha,e_j]$, where $\alpha\in\mathbb{R}^{d}$ embeds the type, parameter and magnitude of the intervention.
The attention output, with $\rho_\tau$ added again, gives the candidate response $C_{i,\tau}=W(q_{i,\tau},s,I_{\le\tau},G)\in\mathbb{R}^{d}$, which thus depends on what the action is, where it acts and the pre-intervention states around the target.

\subsection{Graph-Aware Locality Gate}
\label{sec:gate}
SCRW decides what to write at each node--time position, yet most positions stay quiet under an action (Section~\ref{sec:pairst-response}), so writing every candidate in full would create spurious changes.
GLG provides \textit{selectivity} by assigning each candidate a continuous weight $g_{i,\tau}=\operatorname{Gate}(s_i,I_{\le\tau},\tau,G)\in(0,1)$, which keeps the candidate response $C_{i,\tau}$ when near one and suppresses it when near zero, while the content and direction of the response always come from $C_{i,\tau}$.
A two-layer perceptron with a sigmoid output sets this weight from where node $i$ lies relative to the target ($e_i$), its pre-intervention state ($s_i$), the magnitude of the action and its phase at $\tau$ (the time relative to onset and the release features), so the gate can stay open while the intervention is active and adapt once it has ended.
Because the weight is learned rather than fixed to decay with hop distance, the gate can also adapt its spatial reach to each system (Figure~\ref{fig:pairst-reach}).
Exact pre-onset zeros come instead from the hard mask introduced next in Section~\ref{sec:invariances}.

\begin{table*}[t]
\caption{Results on six PairST datasets. Best in bold, second-best underlined. Datasets are named without the PairST- prefix. GWNet: Graph WaveNet. Avg.\ rank: mean rank over the 18 dataset and metric rows (lower is better).}
\label{tab:main}
\centering
\footnotesize
\setlength{\tabcolsep}{1.4pt}
\begin{tabular*}{\textwidth}{@{}@{\extracolsep{\fill}}cccccccccc@{}}
\toprule[1.5pt]
\textbf{Dataset} & \textbf{Metric} & \textbf{STGCN} & \textbf{GWNet} & \textbf{AGCRN} & \textbf{ConFormer} & \textbf{IGSTGNN} & \textbf{STAEformer} & \textbf{STID} & \textbf{\method{} (ours)} \\
\midrule
\multirow{3}{*}{\textsc{T16}} & {\scriptsize MAE} & \pmcell{1.334}{0.005} & \pmcell{1.383}{0.018} & \pmcell{1.492}{0.030} & \secondpm{1.279}{0.007} & \pmcell{1.293}{0.014} & \pmcell{1.316}{0.019} & \pmcell{1.305}{0.008} & \bestpm{1.249}{0.010} \\ %
 & {\scriptsize AMAE} & \pmcell{10.727}{0.552} & \pmcell{10.236}{0.480} & \pmcell{20.002}{0.880} & \secondpm{7.449}{0.208} & \pmcell{8.925}{0.019} & \pmcell{8.833}{0.340} & \pmcell{8.805}{0.757} & \bestpm{6.164}{0.539} \\ %
 & {\scriptsize QMAE} & \pmcell{0.233}{0.014} & \secondpm{0.175}{0.005} & \pmcell{0.267}{0.051} & \pmcell{0.347}{0.023} & \pmcell{0.177}{0.009} & \pmcell{0.295}{0.023} & \pmcell{0.334}{0.020} & \bestpm{0.162}{0.002} \\ %
\midrule
\multirow{3}{*}{\textsc{T196}} & {\scriptsize MAE} & \pmcell{0.716}{0.004} & \pmcell{0.711}{0.010} & \pmcell{0.724}{0.008} & \pmcell{0.805}{0.035} & \secondpm{0.681}{0.014} & \pmcell{0.690}{0.024} & \pmcell{0.819}{0.009} & \bestpm{0.663}{0.005} \\ %
 & {\scriptsize AMAE} & \pmcell{10.508}{0.046} & \secondpm{8.196}{0.618} & \pmcell{11.918}{0.878} & \pmcell{8.745}{0.564} & \pmcell{8.407}{1.631} & \pmcell{8.539}{2.600} & \pmcell{9.274}{0.071} & \bestpm{4.440}{0.156} \\ %
 & {\scriptsize QMAE} & \pmcell{0.156}{0.001} & \pmcell{0.116}{0.006} & \pmcell{0.163}{0.035} & \pmcell{0.454}{0.047} & \bestpm{0.097}{0.008} & \pmcell{0.130}{0.028} & \pmcell{0.430}{0.029} & \secondpm{0.098}{0.004} \\ %
\midrule
\multirow{3}{*}{\textsc{P59}} & {\scriptsize MAE} & \pmcell{3.088}{0.093} & \pmcell{3.075}{0.030} & \pmcell{2.976}{0.107} & \pmcell{3.194}{0.206} & \pmcell{3.000}{0.186} & \secondpm{2.718}{0.121} & \pmcell{3.325}{0.041} & \bestpm{2.570}{0.065} \\ %
 & {\scriptsize AMAE} & \pmcell{19.017}{0.988} & \pmcell{18.105}{0.552} & \pmcell{21.336}{0.104} & \pmcell{17.855}{0.178} & \bestpm{17.326}{0.435} & \pmcell{18.814}{0.966} & \secondpm{17.698}{0.150} & \pmcell{18.059}{0.059} \\ %
 & {\scriptsize QMAE} & \pmcell{0.598}{0.026} & \pmcell{0.507}{0.058} & \secondpm{0.446}{0.050} & \pmcell{1.126}{0.231} & \pmcell{0.601}{0.075} & \pmcell{0.503}{0.086} & \pmcell{1.437}{0.005} & \bestpm{0.151}{0.038} \\ %
\midrule
\multirow{3}{*}{\textsc{P186}} & {\scriptsize MAE} & \pmcell{5.390}{0.036} & \pmcell{5.665}{0.020} & \pmcell{5.807}{0.060} & \pmcell{5.416}{0.020} & \pmcell{5.693}{0.059} & \secondpm{5.341}{0.091} & \pmcell{5.504}{0.044} & \bestpm{5.293}{0.005} \\ %
 & {\scriptsize AMAE} & \pmcell{15.752}{1.024} & \bestpm{14.585}{0.402} & \pmcell{19.540}{0.289} & \secondpm{15.054}{0.115} & \pmcell{15.499}{0.411} & \pmcell{15.603}{0.141} & \pmcell{15.216}{0.352} & \pmcell{15.604}{0.061} \\ %
 & {\scriptsize QMAE} & \pmcell{0.681}{0.056} & \pmcell{0.352}{0.024} & \pmcell{1.037}{0.061} & \pmcell{1.350}{0.138} & \pmcell{0.346}{0.066} & \secondpm{0.302}{0.073} & \pmcell{1.264}{0.084} & \bestpm{0.098}{0.006} \\ %
\midrule
\multirow{3}{*}{\textsc{W119}} & {\scriptsize MAE} & \pmcell{11.281}{0.303} & \bestpm{10.921}{0.190} & \pmcell{11.749}{0.058} & \pmcell{12.393}{0.231} & \secondpm{11.068}{0.280} & \pmcell{11.747}{0.342} & \pmcell{12.147}{0.121} & \pmcell{11.746}{0.202} \\ %
 & {\scriptsize AMAE} & \pmcell{26.745}{0.724} & \secondpm{23.590}{0.527} & \pmcell{25.731}{8.087} & \pmcell{33.159}{0.082} & \pmcell{24.466}{1.243} & \bestpm{21.352}{0.432} & \pmcell{33.845}{0.342} & \pmcell{34.348}{2.185} \\ %
 & {\scriptsize QMAE} & \pmcell{1.925}{0.247} & \secondpm{1.145}{0.023} & \pmcell{2.168}{0.342} & \pmcell{2.590}{0.198} & \pmcell{1.389}{0.051} & \pmcell{1.948}{0.461} & \pmcell{2.644}{0.031} & \bestpm{0.393}{0.306} \\ %
\midrule
\multirow{3}{*}{\textsc{W40}} & {\scriptsize MAE} & \secondpm{0.550}{0.040} & \pmcell{0.603}{0.049} & \pmcell{0.567}{0.049} & \pmcell{0.607}{0.010} & \pmcell{0.584}{0.009} & \pmcell{0.568}{0.034} & \pmcell{0.635}{0.024} & \bestpm{0.489}{0.046} \\ %
 & {\scriptsize AMAE} & \secondpm{5.635}{0.845} & \pmcell{5.949}{1.580} & \pmcell{6.957}{1.994} & \pmcell{6.853}{0.020} & \pmcell{5.912}{0.991} & \bestpm{1.962}{0.124} & \pmcell{7.376}{0.115} & \pmcell{6.444}{1.535} \\ %
 & {\scriptsize QMAE} & \pmcell{0.075}{0.040} & \pmcell{0.108}{0.034} & \secondpm{0.075}{0.028} & \pmcell{0.156}{0.015} & \pmcell{0.092}{0.037} & \pmcell{0.229}{0.016} & \pmcell{0.151}{0.021} & \bestpm{0.017}{0.022} \\ %
\midrule
\multicolumn{2}{@{}c}{\textbf{Avg.\ rank}} & 4.78 & 3.83 & 5.78 & 5.78 & \underline{3.44} & 3.89 & 6.22 & \textbf{2.28} \\ %
\bottomrule[1.5pt]
\end{tabular*}
\end{table*}

\subsection{Constrained Write}
\label{sec:invariances}
The response state combines the candidate response and the gate under hard constraints that determine whether a write is permitted, $\Delta Z_{i,\tau}=\mathbf{1}[\tau\ge\tau_{\mathrm{on}}]\,g_{i,\tau}\,C_{i,\tau}$.
The indicator blocks writes before the onset, and a hard null rule sets the candidate response exactly to zero when the record is empty.

Together with the prefix-causal encoding of Section~\ref{sec:writing}, these constraints give three structural properties, listed in the constrained write of Figure~\ref{fig:framework}b.
Writing $\Delta Z^{I}$ for the response state under record $I$, they are
\textbf{(i)} \emph{null identity}, $I=\varnothing\Rightarrow\Delta Z^{I}=\mathbf 0$, i.e., an empty record writes nothing,
\textbf{(ii)} \emph{pre-onset invariance}, $\tau<\tau_{\mathrm{on}}\Rightarrow\Delta Z^{I}_{i,\tau}=\mathbf 0$, i.e., nothing is written before the onset, and
\textbf{(iii)} \emph{prefix consistency}, $I\equiv_{\tau}I'\Rightarrow\Delta Z^{I}_{:,\le\tau}=\Delta Z^{I'}_{:,\le\tau}$, i.e., two records with the same executed prefix write the same response up to $\tau$.
Because $W_{\mathrm{out}}$ is bias-free, Eq.~\ref{eq:reference-response} carries these properties to the forecast, so an empty record gives $\hXI=\hXz$ exactly.

\subsection{Training Objective}
\label{sec:objective}
Having defined how responses are written, we train the model to predict the reference and intervened futures, capture action-induced changes, and limit spurious responses in quiet regions.
The base predictor and the response path are trained jointly from scratch, and paired targets enter only the objective, never the model inputs.
$\mathcal{L}_{\mathrm{pred}}$ sums the mean absolute error (MAE) of the reference and intervened futures, which supervises both outputs.
$\mathcal{L}_{\mathrm{resp}}$ trains the affected-region response by combining the response MAE on affected positions with a one-third-weighted medium stratum, $a_m/2\le|\Delta X|<a_m$.
Each training term is first averaged within a pair.

$\mathcal{L}_{\mathrm{resp}}$ covers only affected and medium positions, so the response in the quiet region, where most positions lie, is supervised only indirectly through $\mathcal{L}_{\mathrm{pred}}$.
To supervise \textit{selectivity} directly, we measure the magnitude of the predicted response there,
\begin{equation}
E_{\mathrm{quiet}}=\frac{1}{|\Omega^{\mathrm{quiet}}_m|}\sum\nolimits_{u\in\Omega^{\mathrm{quiet}}_m}|\Delta\hat{X}_u|.
\label{eq:quiet-magnitude}
\end{equation}
A hinge penalizes it only beyond the tolerance $\epsilon$, the 90th percentile over training pairs of the mean true $|\Delta X|$ in the quiet region, so predicted quiet changes may reach the simulator's own quiet magnitude rather than being forced to zero.
The overall loss function is obtained by combining these terms,
\begin{equation}
\mathcal{L}=\mathcal{L}_{\mathrm{pred}}+\mathcal{L}_{\mathrm{resp}}+\lambda_q\,[E_{\mathrm{quiet}}-\epsilon]_+,
\label{eq:native-loss}
\end{equation}
where $\lambda_q$ weights the quiet-region hinge.
Finally, no consistency penalty is needed because Eq.~\ref{eq:reference-response} gives $\Delta\hat{\mathbf{X}}=\hXI-\hXz=\boldsymbol{\sigma}\odot(W_{\mathrm{out}}\Delta Z)$ by construction.

\section{Experiments}
\label{sec:exp}

In this section, we evaluate \method{} through four research questions (\textbf{RQ}s):
\begin{itemize}[leftmargin=*,topsep=2pt,itemsep=2pt,parsep=0pt]
\item \textbf{RQ1.} How does \method{} compare with existing ST forecasters under actions at new targets?
\item \textbf{RQ2.} Does \method{} forecast affected regions accurately without spurious changes in quiet regions?
\item \textbf{RQ3.} Do its gains extend beyond the intervention target?
\item \textbf{RQ4.} How do the response-path components affect accuracy and selectivity?
\end{itemize}

\subsection{Experimental Setup}
\label{sec:exp-setup}

\textbf{Baselines.}
For comparison, we select seven existing ST forecasters, i.e., STGCN~\citep{yu2018stgcn}, Graph WaveNet~\citep{wu2019graphwavenet}, AGCRN~\citep{bai2020agcrn}, ConFormer~\citep{wang2026conformer}, IGSTGNN~\citep{fan2026igstgnn}, STAEformer~\citep{liu2023staeformer} and STID~\citep{shao2022stid}.
To compare all methods on the same outputs, we adapt each baseline to read the intervention record as nine additional input channels at every node and history step, which mark the target and encode the timing and settings of the action, and to output a reference forecast $\hXz$ and an explicit response $\Delta\hat{\mathbf{X}}$, whose sum gives $\hXI$ (Appendix~\ref{app:baselines}).

\textbf{Evaluation Metrics.}
For any set $A$ of scored positions, pooled across test pairs, define
\begin{equation}
\operatorname{Err}_{A}(\hat{Y},Y)
=\frac{1}{|A|}\sum\nolimits_{u\in A}|\hat{Y}_u-Y_u|.
\label{eq:region-error}
\end{equation}
The core errors are
\begin{equation}
\begin{aligned}
\mathrm{MAE}&=\operatorname{Err}_{\Omega}(\hXI,\XI),\\
\mathrm{AMAE}&=\operatorname{Err}_{\Omega^{\mathrm{aff}}_m}(\hXI,\XI),\\
\mathrm{QMAE}&=\operatorname{Err}_{\Omega^{\mathrm{quiet}}_m}(\Delta\hat{\mathbf{X}},\Delta\mathbf{X}).
\end{aligned}
\label{eq:evaluation-metrics}
\end{equation}
MAE is the primary task metric.
Because affected positions are sparse (Section~\ref{sec:pairst-response}), MAE alone can hide both a missed response and a fabricated one, so AMAE scores the intervened forecast where the simulator responds and QMAE scores the response error in the quiet region.

Note that AMAE combines errors at the intervention target $i^\star$ with errors at affected non-target positions, so we also decompose it into
\begin{equation}
\begin{aligned}
\mathrm{T\text{-}AMAE}&=\operatorname{Err}_{\{u\in\Omega^{\mathrm{aff}}_m:\,i=i^\star\}}(\hXI,\XI),\\
\mathrm{NT\text{-}AMAE}&=\operatorname{Err}_{\{u\in\Omega^{\mathrm{aff}}_m:\,i\neq i^\star\}}(\hXI,\XI).
\end{aligned}
\label{eq:propagation-metrics}
\end{equation}
For all metrics, lower values indicate better performance.

\textbf{Evaluation Protocol.}
Each learned method is executed three times on the same fixed data split, and we report the mean and standard deviation of each metric.
Every selection uses validation data.
All tables score forecasts issued one hour before the intervention onset.

\textbf{Implementation Details.}
All learned methods are implemented in PyTorch and trained with AdamW at a learning rate of $10^{-3}$ and a batch size of 16.
Appendix~\ref{app:implementation} gives the baseline implementations and adaptations, the architecture and per-dataset configuration of \method{}, and the training and selection protocol.
\begin{figure}[t]
\centering
\includegraphics[width=\columnwidth]{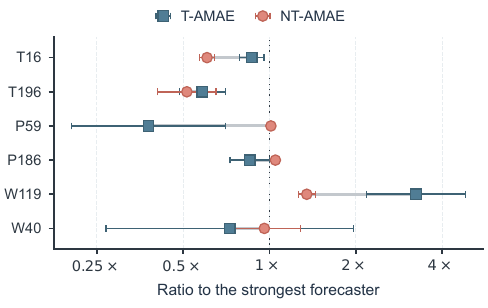}%
\vspace{-1em}
\caption{Errors at and beyond the target, shown as ratios of \method{} to the strongest existing forecaster (below $1\times$ is better). Gray segments connect the two regional estimates for each dataset, and bars propagate standard deviations on the log-ratio scale.}
\label{fig:propagation-error}
\end{figure}
\subsection{Model Comparison (RQ1)}
\label{sec:exp-overall}
To address \textbf{RQ1}, the MAE entries of Table~\ref{tab:main} compare the overall intervened-future accuracy of the existing forecasters and \method{} on the six datasets.
\method{} attains the lowest MAE on five of the six datasets, 0.9\% to 11.0\% below the most accurate existing forecaster on each, and the best average rank.
On PairST-P186, the margin lies within the runner-up's run-to-run spread.
On PairST-W119, it ranks fourth, and two of its three runs predict almost no response.

\begin{figure*}[!t]
\centering
\includegraphics[width=\textwidth]{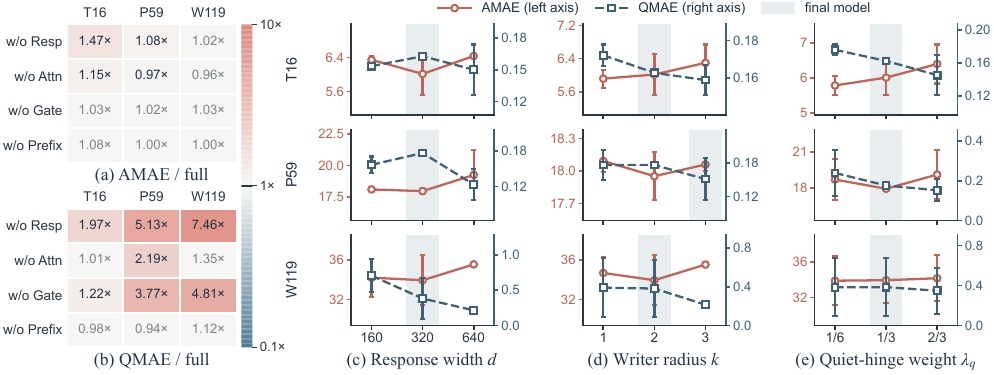}%
\vspace{-1em}
\caption{Component variants and hyperparameter sensitivity. (a, b) Ratios of mean test AMAE and QMAE to the full model, where values above $1\times$ are worse. An entry is unshaded when its absolute mean difference is no larger than the larger run-to-run standard deviation. (c--e) Test AMAE (circles, left axis) and QMAE (squares, right axis) on each dataset, in the units of Table~\ref{tab:main}. Bars show the standard deviation over runs.}
\label{fig:mechanism}
\vspace{-1em}
\end{figure*}

\subsection{Response Evaluation (RQ2)}
\label{sec:exp-response}
To address \textbf{RQ2}, the AMAE and QMAE entries of Table~\ref{tab:main} compare affected-region accuracy and quiet-region selectivity.
Because predicting no response already gives a low QMAE, the two errors are read together.
On traffic, \method{} attains the lowest AMAE on both datasets, 17\% and 46\% below the most accurate existing forecaster, while its QMAE is the lowest on PairST-T16 and within 2\% of the lowest on PairST-T196.
On power, it lowers QMAE by about two thirds relative to every existing forecaster, while its AMAE stays within 7\% of the best.
On water, its QMAE is again the lowest, but its AMAE is higher than that of most existing forecasters, so the low quiet-region error there partly reflects a weaker response rather than better selectivity.
Within each system, the second dataset reproduces the pattern of the first.

\subsection{Response Propagation (RQ3)}
\label{sec:exp-propagation}
To examine whether the gains extend beyond the intervention target, Figure~\ref{fig:propagation-error} compares the T-AMAE and NT-AMAE of \method{} with those of the strongest existing forecaster, i.e., the one with the lowest validation MAE on each dataset.
On the traffic datasets, T-AMAE decreases by 13\% and 42\%, and NT-AMAE by 39\% and 48\%, on PairST-T16 and PairST-T196, respectively.
On the power datasets, the gain concentrates at the target.
\method{} reduces T-AMAE by 62\% on PairST-P59 and by 15\% on PairST-P186, while NT-AMAE differs from the strongest forecaster by at most 5\%.
Water marks the current boundary: PairST-W40 shows lower mean ratios with large run-to-run variation, whereas PairST-W119 is worse at both target and affected non-target positions.
Thus, on both traffic datasets, \method{}'s forecasting gains extend beyond the intervention target.

\subsection{Ablation Study (RQ4)}
\label{sec:exp-mechanism}
To address \textbf{RQ4}, we compare four variants on PairST-T16 and PairST-P59, keeping the base predictor, data, objective and training rule fixed (Figure~\ref{fig:mechanism}a,b): \textbf{w/o Resp} lets the base predictor read the intervention directly instead of using a separate response path, \textbf{w/o Attn} replaces the attention writer with a parameter-matched node--time MLP, \textbf{w/o Gate} removes the locality gate, and \textbf{w/o Prefix} exposes the full intervention record instead of its executed prefix.

Response separation provides the largest benefit, as w/o Resp roughly doubles QMAE on PairST-T16 and increases it more than fivefold on PairST-P59.
Removing the gate raises QMAE by 22\% and nearly fourfold on the two datasets, respectively, while leaving AMAE essentially unchanged, whereas replacing the attention writer raises PairST-T16 AMAE by 15\% and more than doubles PairST-P59 QMAE.
Exposing the full record does not reduce AMAE or QMAE beyond run-to-run variation, so prefix-causal encoding provides prefix consistency (Section~\ref{sec:invariances}) at no cost in accuracy.
Overall, response separation contributes most, while the attention writer and the gate preserve affected-region accuracy and quiet-region selectivity.

\subsection{Hyperparameter Study}
\label{sec:exp-hp}
We vary the response width $d$, the writer radius $k$ and the quiet-hinge weight $\lambda_q$ one at a time on PairST-T16, PairST-P59 and PairST-W119.
These control the capacity, neighbourhood context and quiet-region emphasis of the response path, respectively (Figure~\ref{fig:mechanism}c--e).
The MAE of the intervened future changes by at most 5.5\% across the tested ranges, whereas AMAE and QMAE vary more and expose a trade-off between affected-region and quiet-region error.
The middle response width gives the lowest AMAE on all three datasets.
Increasing $k$ lowers QMAE on PairST-T16 and PairST-P59 without moving AMAE beyond run-to-run variation.
Increasing $\lambda_q$ from $1/6$ to $2/3$ lowers QMAE on these two datasets by 17\% and 37\%, respectively, although AMAE rises on PairST-T16.
On PairST-W119, the largest width and radius suppress the response almost entirely, so their lower QMAE does not indicate improved selectivity.

\section{Conclusion}
\label{sec:conclusion}

We studied action-conditioned ST forecasting, which asks how a networked system will evolve under a specified action.
We constructed PairST, whose same-state paired futures on six datasets across traffic, power transmission and water distribution supervise and evaluate the response directly.
Building on these pairs, \method{} forecasts the intervened future by writing an explicit response onto an intervention-independent reference forecast.
At new action targets, \method{} gave a more accurate intervened forecast than the compared forecasters on five of six datasets.
Its affected-region advantage was largest on traffic, where it extended beyond the intervention target, and on power it lowered quiet-region response error by about two thirds with affected-region error within 7\% of the best.
PairST currently covers one action type per system and a single action at one target in each scenario.
In the future, extending PairST to more action types, repeated actions and simultaneous actions at multiple locations is a promising research direction.

\clearpage
\bibliography{ref}
\bibliographystyle{icml2025}

\newpage
\appendix
\renewcommand{\thefigure}{\Alph{section}.\arabic{figure}}
\renewcommand{\thetable}{\Alph{section}.\arabic{table}}
\renewcommand{\theHfigure}{appendix.\Alph{section}.\arabic{figure}}
\renewcommand{\theHtable}{appendix.\Alph{section}.\arabic{table}}
\setcounter{figure}{0}
\setcounter{table}{0}
\counterwithin*{figure}{section}
\counterwithin*{table}{section}
\onecolumn
\renewenvironment{table*}[1][]{\begin{table}[H]}{\end{table}}
\raggedbottom

\section{\texorpdfstring{PairST}{PairST} Dataset Details}
\label{app:pairst-details}
This section gives the generation, split and threshold settings of the PairST datasets and characterizes their simulator response, computed on training pairs unless a split is named.

\subsection{Generation Details}
\label{app:thresholds}
\textbf{Traffic.}
Both datasets run in CityFlow~\citep{zhang2019cityflow} and record the mean vehicle count on the incoming lanes of each intersection in five-minute bins.
PairST-T16 uses the anonymized CoLight Hangzhou $4\times4$ assets~\citep{wei2019colight} and PairST-T196 the CoLight New York $28\times7$ assets, and both are source assets rather than geographically faithful city maps.
Demand is synthetic on both, with departures following a shared daily profile and routes resampled from each asset's route pool.

\textbf{Power.}
PairST-P59 runs the small NeurIPS-2020 Track-1 environment of Grid2Op~\citep{donnot2020grid2op,marot2021l2rpn}, with 36 substations and 59 lines, and PairST-P186 the WCCI-2022 environment, with 118 substations and 186 lines, both with the opponent disabled and the maintenance schedule retained.
Power flows are solved with PandaPower on PairST-P59 and LightSim2Grid on PairST-P186.
Both datasets record line loading, the flow of each line as a percentage of its thermal limit, at the native five-minute step, and each day is one load and generation scenario of the environment.

\textbf{Water.}
Both datasets run EPANET~\citep{rossman2020epanet} through WNTR~\citep{klise2017wntr} with pressure-driven demand and record signed link flow in litres per second (L/s) in fifteen-minute bins.
Each day multiplies the base demand of every junction by the exponential of a graph-smoothed spatial field and a temporal AR(1) process, with standard deviations $0.15$ and $0.10$ and temporal correlation $0.95$, clipped to $[0.5,1.6]$.
The spatial standard deviation is taken over the positive-demand junctions on PairST-W40 and over all junctions on PairST-W119.

\begin{figure}[!t]
\centering
\includegraphics[width=0.88\textwidth]{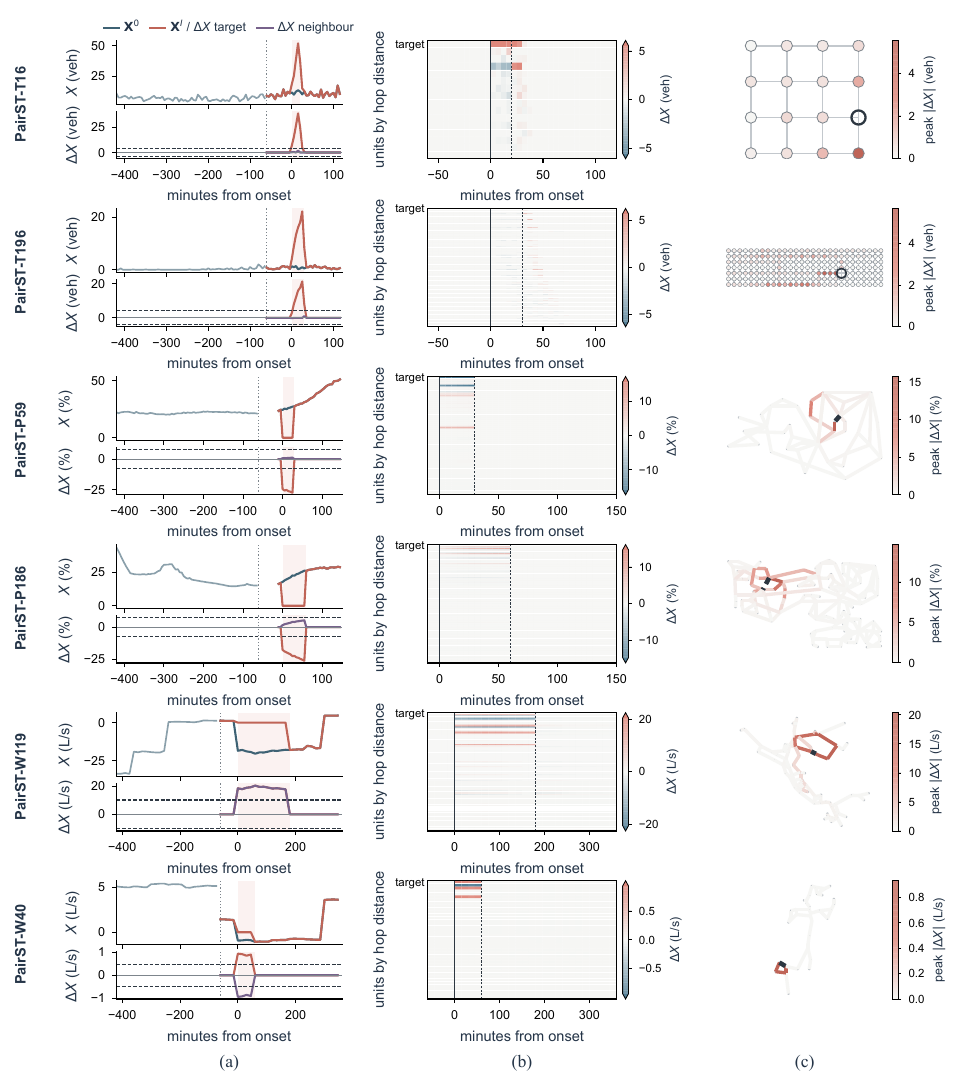}
\caption{Network layouts and one training pair per dataset, chosen by the median rule above. (a) Target trajectory without ($\Xz$) and with ($\XI$) the intervention, after the history tail ending at the forecast time (dotted), and $\Delta X$ at the target and one neighbour, with $\pm a_m$ dashed. (b) $\Delta X$ of every unit by hop distance, with the colour range set by the non-target units and the intervention start (solid) and end (dotted). (c) Peak $|\Delta X|$ on the schematic layout, with the target outlined or black and excluded from the colour scale. Shading marks the intervention.}
\label{fig:pairst-examples}
\end{figure}

\textbf{Graphs.}
On traffic, each node is an intersection, and a directed edge runs from one intersection to the next when a road leaves the first and enters the second.
On power, each node is a transmission line, and two lines are connected when they share a substation.
On water, each node is a link, and two links are connected when they share a junction, tank or reservoir, so the power and water graphs are undirected.
Hop distances follow these edges, and the direction in the target-relative encoding is the compass direction between the planar positions of two nodes.
Figure~\ref{fig:pairst-examples}c draws each graph on the schematic layout of its asset, together with one training pair per dataset.
The pair is the one closest to the dataset's training medians of the target peak, the non-target peak and the number of affected non-target units, and the neighbour is the lowest-indexed unit one hop from the target.

\textbf{History and Forecast Windows.}
In every dataset the first simulated day supplies the 24-hour history, and every intervention starts on the second day.
Every onset lies at least 8 hours into the second day, so a forecast issued up to 8 hours before the onset, the longest lead the generators use, still has the full 24-hour history as input.
The history ends at the forecast time, and the output window starts at or after it.

\textbf{Onsets and Durations.}
The four traffic onsets fall at the morning demand peak, two hours after it, and one hour before and after the evening peak, and the power and water onsets are spread over the day, every four and every three hours.
Each location carries eight records, four onsets crossed with two durations, and the same grid applies in every split (Table~\ref{tab:grid}).
On traffic, the eight records of a location set each of the four signal phases equally often.

\begin{table}[H]
\caption{Record grid of each dataset. Onsets: hours after the start of the second simulated day. Output window: relative to the onset.}
\label{tab:grid}
\centering
\footnotesize
\begin{tabular}{@{}lcccc@{}}
\toprule[1.5pt]
\textbf{Dataset} & \textbf{Bin} & \textbf{Onsets (h)} & \textbf{Durations} & \textbf{Output window} \\
\midrule
PairST-T16 & 5 min & 8, 10, 16, 18 & 20, 25 min & $-1$ h to $+2$ h \\
PairST-T196 & 5 min & 8, 10, 16, 18 & 30, 40 min & $-1$ h to $+2$ h \\
PairST-P59 & 5 min & 8, 12, 16, 20 & 30, 60 min & $-10$ min to $+2.5$ h \\
PairST-P186 & 5 min & 8, 12, 16, 20 & 30, 60 min & $-10$ min to $+2.5$ h \\
PairST-W119 & 15 min & 8, 11, 14, 17 & 1, 3 h & $-1$ h to $+6$ h \\
PairST-W40 & 15 min & 8, 11, 14, 17 & 1, 3 h & $-1$ h to $+6$ h \\
\bottomrule[1.5pt]
\end{tabular}
\end{table}

\textbf{Response Thresholds.}
The traffic threshold $a_m=4$ vehicles is shared by both traffic datasets and was fixed without reference to their responses.

\subsection{Splits and Qualification}
Eligible targets are, for traffic, the signalized intersections whose signal plan carries these four phases, for power the lines whose disconnection does not split the grid and yields a valid outcome, chiefly one without game over, on at least 95\% of calibration scenarios, and for water the initially open pipes whose closure does not disconnect the network.
Each dataset ranks its eligible targets by the response size of a calibration screen and selects $\min(24,\text{eligible})$ locations stratified by that size, drawn by a seeded rule on traffic and water and taken as the lower median of each stratum on power.
A seeded assignment within strata then splits them into training, validation and test in proportions of about $2/3$, $1/6$ and $1/6$, so that every split spans the range of response sizes.
PairST-T16 and PairST-W40 have fewer than 24 eligible targets and use all of them.
Days are drawn by each generator's seeded rule, and no day belongs to two splits.
A power day is accepted only when its two no-action replays agree and every record of its split is valid on it.
Table~\ref{tab:data} gives the final counts.

Before any model is trained, each dataset must show exact paired replay, an acceptance rate of at least $0.95$, a response rate of at least $0.50$, a quiet-position share of at least $0.25$, and no training location or day in validation or test.
Response rate is the share of accepted pairs with an active response outside the target.

\subsection{Characteristics of Action-Induced Changes}
Beyond the single pairs of Figure~\ref{fig:pairst-examples}, we characterize action-induced changes in PairST through their magnitude and sparsity, spatial extent, temporal evolution and distributions across data splits.

\begin{figure}[t]
\centering
\includegraphics[width=\textwidth]{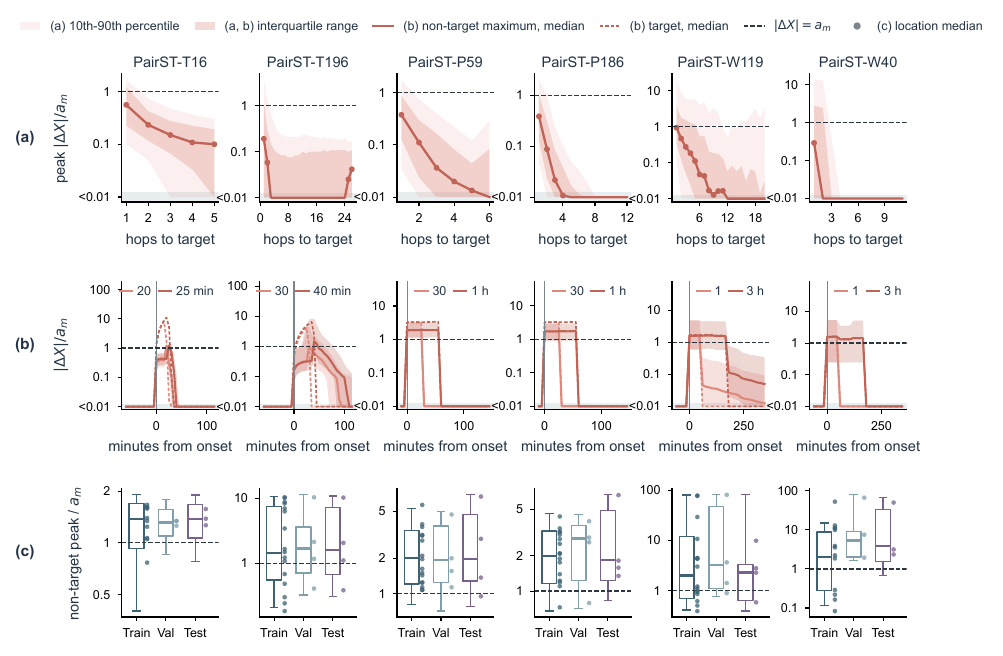}
\caption{Response characteristics of the PairST datasets, relative to $a_m$. (a) Window peak $|\Delta X|$ of non-target units by hop distance: median, interquartile range and 10th to 90th percentile, at hops reached from at least half of the training target locations. (b) Largest non-target $|\Delta X|$ (solid, median and interquartile range) and target $|\Delta X|$ (dashed, median) at each step, per duration, with the onset marked. (c) Non-target peak by split, as boxes with 10th to 90th percentile whiskers and one dot per location median. Dashed lines mark $a_m$, and values below $0.01\,a_m$ lie in the grey band.}
\label{fig:pairst-stats}
\end{figure}

\textbf{Response Size and Sparsity.}
Table~\ref{tab:pairst-response} reports the response size and sparsity of each dataset in its native unit, and PairST-W119 has both the largest share of affected positions and the heaviest tail of non-target peaks.
\begin{table}[!tb]
\caption{Response characteristics of the PairST datasets, with shares in percent. $|X^0|$: median no-intervention magnitude over day and time bins. Non-target peak: largest $|\Delta X|$ outside the target in a pair's output window, median [interquartile range]. Affected, Quiet: shares of output positions, target included. Units aff.: median share of non-target units affected at some step. Not recovered: pairs with a non-quiet unit in the last 30 minutes of the window, short / long duration.}
\label{tab:pairst-response}
\centering
\footnotesize
\setlength{\tabcolsep}{4pt}
\begin{tabular}{@{}lccccccc@{}}
\toprule[1.5pt]
\textbf{Dataset} & \textbf{Unit} & $\boldsymbol{|X^0|}$ & \textbf{Non-target peak} & \textbf{Affected} & \textbf{Quiet} & \textbf{Units aff.} & \textbf{Not recovered} \\
\midrule
PairST-T16 & veh & 5.1 & 5.5 [3.7, 6.8] & 1.3 & 97.7 & 6.7 & 0.7 / 0.9 \\
PairST-T196 & veh & 1.0 & 5.8 [2.2, 30.0] & 0.5 & 99.1 & 1.0 & 15.7 / 27.8 \\
\midrule
PairST-P59 & \% & 26.4 & 15.6 [9.4, 26.7] & 1.8 & 96.5 & 3.4 & 0.9 / 1.0 \\
PairST-P186 & \% & 27.7 & 14.8 [8.5, 24.2] & 0.6 & 99.0 & 1.1 & 0.9 / 1.1 \\
\midrule
PairST-W119 & L/s & 14.2 & 20.4 [7.1, 122] & 7.3 & 90.0 & 8.5 & 17.9 / 23.2 \\
PairST-W40 & L/s & 1.3 & 0.93 [0.13, 4.1] & 2.3 & 97.4 & 7.7 & 0.0 / 0.0 \\
\bottomrule[1.5pt]
\end{tabular}
\end{table}

\textbf{Spatial Extent.}
The response of a non-target unit shrinks with its hop distance to the target, and at no distance on any dataset does the median unit reach $a_m$, so an affected non-target unit is the exception at every distance (Figure~\ref{fig:pairst-stats}a).

\textbf{Temporal Evolution.}
On most datasets the response begins at the onset and returns to quiet within the output window, and only on PairST-W119 and PairST-T196 does part of it outlast the intervention (Figure~\ref{fig:pairst-stats}b and Not recovered in Table~\ref{tab:pairst-response}).

\textbf{Split Comparability.}
On every dataset the non-target peaks of the training, validation and test splits overlap, although each validation and test split has only three or four target locations, which set its distribution (Figure~\ref{fig:pairst-stats}c).

\section{Experimental Implementation}
\label{app:implementation}
This section gives the implementation and training settings behind every result table.

\subsection{Baselines and the Shared Intervention Interface}
\label{app:baselines}
We compare seven ST forecasting baselines and adapt them to a shared intervention interface.
Each baseline chooses its input length from 1, 4 and 24 hours by the validation MAE of $\XI$ in the first run.
Except for STAEformer, which keeps its released width, its width is selected against a budget of 11.8 million trainable parameters, within $\pm10\%$ where its width range permits and otherwise at the closest width.
\begin{itemize}[leftmargin=*, itemsep=2pt, topsep=2pt, parsep=0pt, partopsep=0pt]
\item
\textbf{STGCN~\citep{yu2018stgcn}.}
STGCN alternates gated temporal convolutions and Chebyshev graph convolutions in two ST-Conv blocks.
We translate the authors' TensorFlow implementation\footnote{\url{https://github.com/VeritasYin/STGCN_IJCAI-18/tree/4ab75ab}} to PyTorch and replace its single-step readout with the shared multi-horizon output interface.
It needs at least 10 input steps, so on water it uses only the 4- and 24-hour inputs.
Its parameter count grows with the input length, so its width (115 or 116 on traffic and power, 170 or 171 on water) matches the budget on average over the feasible input lengths, and single lengths differ from it by up to 158\%.
\item
\textbf{Graph WaveNet~\citep{wu2019graphwavenet}.}
Graph WaveNet combines dilated causal temporal convolutions with graph convolutions over predefined and learned adjacency matrices.
We use the CausalST copy of its PyTorch implementation\footnote{\url{https://github.com/nnzhan/Graph-WaveNet/tree/6b162e8}} and adapt its input and output projections to the shared interface.
Its width is the largest of its range, 128, at 83\% below the budget.
\item
\textbf{AGCRN~\citep{bai2020agcrn}.}
AGCRN combines recurrent units with node-adaptive graph convolutions, learning node-specific parameters and an adaptive adjacency matrix.
We use the official implementation\footnote{\url{https://github.com/LeiBAI/AGCRN/tree/7fbbf2a}} with the shared record inputs and two-output forecasting interface.
Its width of 254 matches the budget, and on PairST-P186 its 24-hour input exceeds the available accelerator memory and is not run.
\item
\textbf{ConFormer~\citep{wang2026conformer}.}
ConFormer combines spatial attention with graph-propagated incident context that modulates layer normalization.
We use the official implementation\footnote{\url{https://github.com/Dreamzz5/ConFormer/tree/743603f}} with the target indicator as its accident flag, a zero regulation flag and the shared record encoding as additional inputs.
Its width of 96, or 97 on water, matches the budget within 2\%.
\item
\textbf{IGSTGNN~\citep{fan2026igstgnn}.}
IGSTGNN fuses incident context with traffic states through incident--sensor attention and models the temporal decay of incident effects.
We use the official end-to-end implementation\footnote{\url{https://github.com/fanlixiang/IGSTGNN/tree/64c6e24}} and encode the action type, parameter and target-relative graph distances through its incident pathway, alongside the shared record channels.
It needs at least 11 input steps, so on water it uses only the 4- and 24-hour inputs, and on PairST-P186 its 24-hour input exceeds the available accelerator memory and is not run.
Its width is the largest of its range, 128, at 71\% to 72\% below the budget.
\item
\textbf{STAEformer~\citep{liu2023staeformer}.}
STAEformer combines temporal and spatial self-attention with an adaptive ST embedding.
We use the official implementation\footnote{\url{https://github.com/XDZhelheim/STAEformer/tree/fc49d39}}, append the shared record channels to its inputs, and retain its released embedding width of 24 on every dataset.
\item
\textbf{STID~\citep{shao2022stid}.}
STID combines node and time identity embeddings with residual MLP blocks.
We use the official implementation\footnote{\url{https://github.com/GestaltCogTeam/STID/tree/e8b313b}}, append the shared record encoding, and expand its output projection to produce reference and response channels over the forecast window.
Its width is the largest of its range, 128, at 82\% to 86\% below the budget.
\end{itemize}

\textbf{Shared Intervention Interface.}
The adapters provide each baseline and \method{} with the same history, graph and intervention record.
Each baseline reads the record as nine shared input channels, appended to the history value at every input position and read through each model's own input projection: the target indicator, the time relative to the onset, an in-window flag, the duration, the action parameter, the magnitude, the lead, the time of day and the day index.
One network computes a reference channel and a response channel from these inputs, giving $\hXI=\hXz+\Delta\hat{\mathbf{X}}$.
An empty record gives exactly zero response, and QMAE scores the explicit response channel against the paired response target.
The reference channel is supervised on $\Xz$ but remains conditioned on the intervention.
Every baseline is trained on the same forecast and response terms as \method{}, with a one-third-weighted penalty on every predicted quiet change in place of the hinge.

\subsection{\texorpdfstring{\method{}}{ActionST} Architecture and Configuration}
\label{app:architecture}\label{app:config}
This section gives the layers and settings of each component in Figure~\ref{fig:framework}.

\textbf{Base Predictor.}
Each node is represented by one token, following STID~\citep{shao2022stid}.
A linear projection of the node's recent values, a node embedding, an embedding of the time of day of the last observed step and a linear embedding of the forecast lead are concatenated into this token.
$B$ residual MLP blocks, each a linear layer, a ReLU, dropout and a second linear layer with a skip connection, refine the token into $h_i$, and a linear readout maps $h_i$ to $\hXz$.
The base predictor uses no graph.
It embeds the recent values, the node, the time of day and the lead with width 128 each, initializes its embeddings with standard deviation 0.1 and uses dropout 0.15.

\textbf{Response Path.}
In SCRW, the writer query $q_{i,\tau}+e_i+\rho_\tau$ passes through a layer normalization before multi-head attention.
The keys and values are obtained from $[s_j,\alpha,e_j]$ by a linear layer, a GELU and a layer normalization.
The attention output plus $\rho_\tau$ passes through a layer normalization and a linear layer to give $C_{i,\tau}$, and the release encoding $\rho_\tau$ is itself a linear layer, a GELU and a linear layer applied to the release features.
GLG is a linear layer, a GELU, a linear layer and a sigmoid applied to its inputs, with 128 hidden units.
The response path has width $d=320$ and dropout 0.1.
The writer's output projection starts at zero, and the gate's output bias starts at $+3$, so $g\approx0.95$ at initialization.

\textbf{Per-Dataset Settings.}
Every dataset has one output feature, so $D'=1$ and the readout $W_{\mathrm{out}}$ in Eq.~\ref{eq:reference-response} is a single row.
The remaining settings are selected per dataset on validation data and are listed in Table~\ref{tab:config}.
Each training batch also contains one example with the empty record, and otherwise the null rule, objective and training rule keep one definition on every dataset.

\begin{table}[H]
\caption{Per-dataset settings of \method{} and the resulting number of trainable parameters.}
\label{tab:config}
\centering
\footnotesize
\begin{tabular}{@{}lcccccc@{}}
\toprule[1.5pt]
\textbf{Setting} & \textbf{PairST-T16} & \textbf{PairST-T196} & \textbf{PairST-P59} & \textbf{PairST-P186} & \textbf{PairST-W119} & \textbf{PairST-W40} \\
\midrule
History length (h) & 4 & 24 & 1 & 24 & 1 & 1 \\
Base predictor blocks $B$ & 2 & 3 & 3 & 3 & 3 & 4 \\
Writer radius $k$ & 2 & 2 & 3 & 2 & 2 & 1 \\
Attention heads & 8 & 16 & 8 & 8 & 8 & 8 \\
Hinge weight $\lambda_q$ & $1/3$ & $1/3$ & $1/3$ & $1/6$ & $1/3$ & $1/3$ \\
Weight decay & $10^{-4}$ & $10^{-4}$ & $10^{-4}$ & $10^{-4}$ & $10^{-2}$ & $10^{-4}$ \\
Parameters (M) & 2.39 & 2.97 & 2.92 & 2.97 & 2.90 & 3.41 \\
\bottomrule[1.5pt]
\end{tabular}
\end{table}

\subsection{Training and Selection Protocol}
During training, the lead between the forecast time and the onset is drawn uniformly from 1, 3 and 6 hours for each example and epoch.
Every learned baseline uses AdamW with learning rate $10^{-3}$, weight decay $10^{-4}$, a cosine schedule and batch size 16, for at most 200 epochs with early stopping after 20 validation evaluations without improvement in the validation MAE of $\XI$.
\method{} uses the same optimizer, learning rate and batch size, with the weight decay of Appendix~\ref{app:config}, and trains on the full training split for 40 epochs under a cosine schedule that decays to zero, without early stopping.
Its checkpoint is the epoch with the lowest validation MAE of $\XI$.
The threshold $\epsilon$ of Eq.~\ref{eq:native-loss} is computed once per dataset, before training, from its training pairs at the one-hour lead.
Oversized baseline batches use gradient accumulation at the same effective batch size.
Batch-normalized models retain the full batch.

For Figure~\ref{fig:propagation-error}, selecting the existing forecaster with the lowest validation MAE yields ConFormer on PairST-T16 and PairST-P186, IGSTGNN on PairST-T196 and PairST-W119, STAEformer on PairST-P59, and AGCRN on PairST-W40.

\end{document}